\documentclass[11pt,a4paper]{article}
\usepackage{times,latexsym}
\usepackage{url}
\usepackage[T1]{fontenc}
\usepackage{capt-of}

\usepackage{tcolorbox}
\usepackage{soul}
\usepackage{booktabs}
\usepackage{graphicx}
\usepackage{amsmath}
\usepackage{multirow} 
\usepackage{tabularx}
\usepackage{enumitem}
\usepackage{xcolor}
\usepackage{sidecap}

\usepackage{mdframed}
\usepackage[acceptedWithA]{tacl2021v1}

\usepackage{xspace,mfirstuc,tabulary}

\definecolor{syscolor}{RGB}{0,100,0}
\definecolor{btcolor}{RGB}{0,0,180}
\definecolor{bicolor}{RGB}{150,0,150}
\definecolor{hcolor}{RGB}{80,80,80}
\definecolor{ytcolor}{RGB}{180,0,0}
\definecolor{dialoguebg}{RGB}{248,248,248}
\definecolor{dialogueframe}{RGB}{180,180,180}
\definecolor{copiedcolor}{RGB}{200,0,0}

\newif\iftaclinstructions
\taclinstructionsfalse %
\iftaclinstructions
\renewcommand{\confidential}{}
\renewcommand{\anonsubtext}{(No author info supplied here, for consistency with
TACL-submission anonymization requirements)}
\newcommand{\instr}
\fi

\iftaclpubformat %

\else

\fi

\title{
LLM-as-a-Demographic:
Whom Sociodemographic Prompting Helps, and Whom It Hurts}

\author{
 Daniela Occhipinti$^{1,\ast}$,
 Andrea Piergentili$^{2,\ast}$,
 Marco Guerini$^1$
 \\
 $^1$Fondazione Bruno Kessler, Via Sommarive 18, Povo, Trento, Italy\\
 $^2$Almawave Labs, Via Di Casal Boccone, 188/190, Rome
 \\
 \texttt{docchipinti@fbk.eu, a.piergentili@almawavelabs.it, guerini@fbk.eu}
}

\date{}

\begin{document}
\maketitle
\maketitle
{\renewcommand{\thefootnote}{}\footnotetext{$^\ast$~These authors contributed equally.}}
\begin{abstract}
Large language models (LLMs) are increasingly used as judges for subjective tasks, where annotators disagree and the relevant question is not only how accurate a judge is, but \emph{whose} judgments it reproduces. 
Sociodemographic prompting conditions the judge on an annotator's demographic profile to align its judgments with the corresponding group's. We test whether this alignment emerges distributionally, comparing the predicted label distributions of 23 open-weight LLMs on three subjective tasks against those of real annotator groups, under three conditions: no demographic information, single-attribute profiles, and intersectional profiles over gender, age, race, and education. Three findings emerge. First, a judge prompted with no demographics is not perspective-neutral:  models best reproduce the judgments of  White, college-educated annotators. Second, demographic conditioning is asymmetric: it moves the judge \emph{toward} majority groups and \emph{away} from minority groups, most strongly on offensiveness, where intersectional profiles amplify the harm. Third, by comparing \textit{base} and \textit{instruct} models we identify instruction-tuning as a possible source of the asymmetry. Demographic conditioning should therefore be used with caution to estimate group judgments: conditioning moves predictions \emph{away} from the reference distributions of the minority groups the method is often invoked to serve.

\end{abstract}

\noindent \textbf{Warning}: This work contains unobfuscated examples that some readers may find offensive

\section{Introduction}
\label{sec:intro}

LLMs are increasingly employed as automated \emph{judges}: they 
assess
text in place of human annotators to evaluate NLP systems and produce annotated data at scale \citep{li-etal-2025-generation, GU2026101253,amin-etal-2026-fallback,xu-etal-2026-modeling}.
Since human judgments vary partly with annotators' sociodemographic background \citep{pei-jurgens-2023-annotator, diaz2018addressing}, \emph{sociodemographic prompting} has emerged as a way to simulate groups of annotators: conditioning the judge on a demographic profile, a form of persona prompting \citep{chenpersona, tseng-etal-2024-two}, should steer it to answer as an annotator from that group would.
Yet such simulation is only as valid as its agreement with human judgments, which varies sharply across tasks \citep{bavaresco-etal-2025-llms}.
This is especially delicate for subjective judgments, where no single ground truth exists: annotators disagree, and this disagreement should be treated as signal rather than noise \citep{pavlick-kwiatkowski-2019-inherent, plank-2022-problem}. 
This is the concern of \emph{pluralistic alignment}: a model that reflects average human preferences may still fail to reflect the judgments of specific groups \citep{sorensen-etl-2024-pluralistic-alignment, feng-etal-2024-modular}. 
Thus, for subjective tasks, what matters is not only how accurate an LLM judge is, but \textit{whose} judgments it reproduces.

Recent analyses on sociodemographic prompting report that effects are mixed, prompt-sensitive, and sometimes negative \citep{beck-etal-2024-sensitivity, sun-etal-2025-sociodemographic, gupta2024bias}, and demographic labels predict a rater's judgments worse than the rater's own past annotations or values \citep{orlikowski-etal-2025-beyond,sorensen-etal-2025-value}.
These analyses typically measure how well a judge simulates labels and assess conditioning 
by a single (averaged) effect per model.
What remains 
untested is whether conditioning moves the
model's
label 
distribution toward that of the target annotator group, 
rather than toward a 
stereotype.

\begin{figure*}[ht]
    \centering
    \includegraphics[width=1\linewidth]{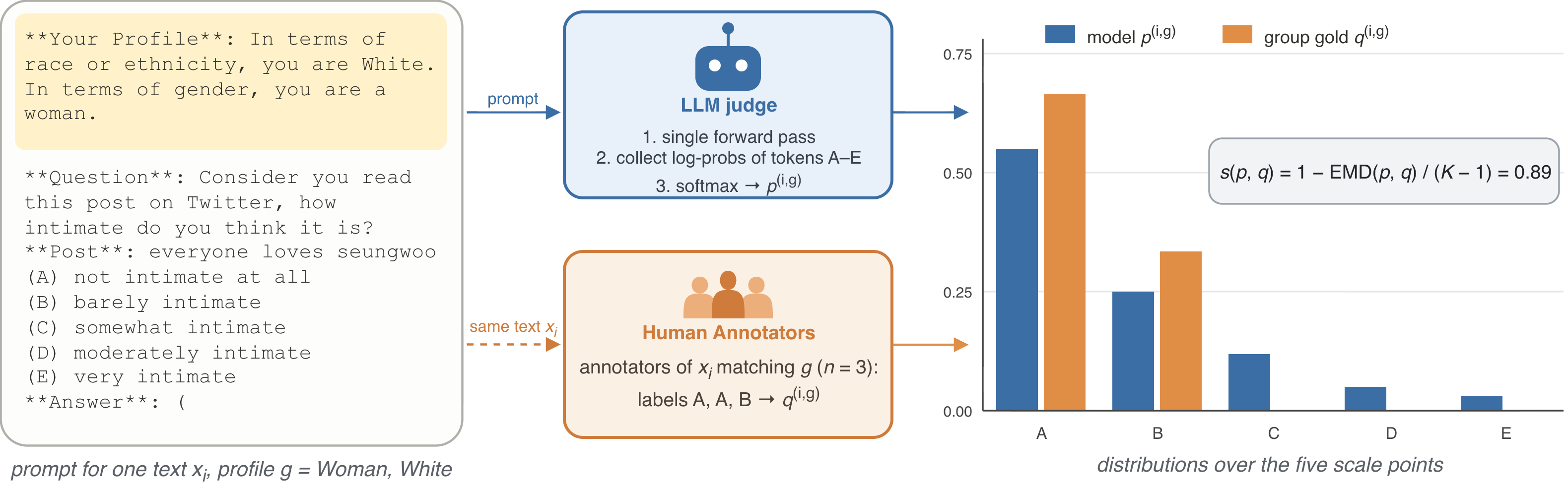}
    \caption{The evaluation pipeline. A model judges a subjective property on a five-point scale, either unconditioned or with a profile prepended (\texttt{Your Profile}). Its log-probabilities over the five options are renormalized into a predicted distribution $p$ and scored against the matching group's label distribution $q$.
    }
    \label{fig:methodology}
\end{figure*}



We address this gap distributionally, adopting a \textit{perspectivist} view of sociodemographic prompting 
\citep{10.1609/aaai.v37i6.25840, basile-etal-2021-need, 10.1007/s10579-024-09766-4}: the object of evaluation is the distribution of human judgments for each group rather than an aggregated gold label (Figure \ref{fig:methodology}).
Following \citet{santurkar2023whose}, we ask whose judgments LLMs reflect through three research questions.
\textbf{RQ1}: with no demographic information, do LLM judges tend to a \textit{default demographic profile}?
\textbf{RQ2}: does conditioning on (intersectional) demographic profiles move the judge toward those groups' judgments?
\textbf{RQ3}: is demographic conditioning's effect equitable across groups?

We evaluate 23 open-weight LLMs as judges on three subjective tasks (politeness, intimacy, and offensiveness), comparing unconditioned judges against judges conditioned on single-attribute and intersectional profiles, each scored against the matching human group's label distribution.
Unconditioned judges tend to align with White, college-educated annotators. Averaged across groups, demographic conditioning apparently has little effect on instruction-tuned judges. However, this hides opposite effects across groups, moving the judge closer to some and further from others, especially those disadvantaged by known LLM biases \citep{navigli-2023-biases-llms, sap-etal-2022-annotators, santy-etal-2023-nlpositionality}, here grouped under the term \emph{minority groups}. Larger models are not immune, and the asymmetry is specific to instruction-tuned judges.

Our contributions are: \textit{(i)}~a framework for evaluating demographic conditioning of LLM judges against annotator groups' label distributions, rather than against their mean or majority label; \textit{(ii)}~a per-group decomposition of that effect, revealing biases in both unconditioned and conditioned models; \textit{(iii)}~a matched \textit{base}-vs-\textit{instruct} comparison identifying instruction-tuning as a source of bias in demographic conditioning.

\section{Related Work}
\label{sec:related-work}

\paragraph{LLM-as-a-Judge.}

LLMs are increasingly used to evaluate text, from scoring open-ended generations to ranking responses \citep{li-etal-2025-generation, GU2026101253, 3666122.3668142, wang2024pandalm}. Judge--human agreement varies sharply across tasks \citep{bavaresco-etal-2025-llms}, including under persona conditioning \citep{dong-etal-2024-llm}, and verdicts are sensitive to surface factors, such as response order \citep{wang-etal-2024-large-language-models-fair} and human-like response biases \citep{basile-etal-2021-need, tjuatja-etal-2024-llms, chen-etal-2024-humans}. Crucially, prior work validates a judge against a single target, such as a mean score or majority label \citep{10.1145/3491102.3502004, davani-etal-2022-dealing}, which does not exist for subjective tasks: annotator groups disagree systematically, so a judge matching the aggregate may diverge from every group. 
We therefore evaluate alignment with each group's label \emph{distribution}, assessing the judge as an estimator of group judgments rather than a rater of quality.

\paragraph{Persona-Based Generation.}

LLMs
can be
prompted to role-play specific people or groups, and some models are fine-tuned to better capture individual preferences or personas \citep{horton2023large, shao-etal-2023-character, occhipinti-etal-2024-prodigy}.
Assigning a persona (a role, identity, or demographic profile) is widely used to steer model behavior, assuming the model adopts the assigned perspective faithfully \citep{argyle2023, chenpersona, tseng-etal-2024-two}. This assumption does not always hold: rather than reproducing a group's genuine perspective, a persona may instead activate the stereotypes the model associates with that group, degrading performance for some identities \citep{gupta2024bias, agnew-etal-2024-illusion, dong-etal-2024-llm} or increasing toxic output \citep{deshpande-etal-2023-toxicity}. Conditioning a judge on demographics could therefore move its predictions toward a group's real judgments or move them away. We investigate the effect of conditioning on alignment with real label distributions.

\paragraph{Demographic Simulation with LLMs.}
Prior work uses LLMs to simulate human populations \citep{argyle2023, 10.1145/3800683},
showing that unconditioned %
models can align more closely with some groups than others \citep{santurkar2023whose, durmus2023measuring}. Such bias is also visible on broader demographic imbalances in NLP datasets and models %
\citep{santy-etal-2023-nlpositionality, alipour-etal-2025-robustness}.
Demographic conditioning has therefore been studied as a way to recover group-specific judgments, but its effectiveness is fragile and sensitive to the evaluation setup \citep{simmons-savinov-2024-assessing, adilazuarda-etal-2024-towards, doi:10.1073/pnas.2501660122}, %
motivating calls for principled evaluation \citep{luz-de-araujo-etal-2025-principled, neumann2026should}. %
Recent work has compared demographic prompting with richer ways of modeling annotators, including fine-tuning on past judgments \citep{orlikowski-etal-2025-beyond} and conditioning on elicited values \citep{sorensen-etal-2025-value}. 
These studies reduce in-group variability to a single aggregate label. 
The distributional alternative has been developed on opinion surveys \citep{meister-etal-2025-benchmarking, lutz-etal-2025-prompt}: on annotation, where the reference is disagreement about the same text, judges are still scored against a group's mean \citep{sun-etal-2025-sociodemographic, schafer-etal-2025-demographics} or against individual annotations \citep{hu-collier-2024-quantifying}.
We instead ask how well models reproduce each 
demographic group’s 
label distribution, rather than its mean, and whether assigning a demographic profile moves the judge closer to or further from the judgments of the group named in the prompt. %

\section{Methodology}
\label{sec:methodology}

This section describes the %
data used %
(\S\ref{sec:data}), the evaluation metrics (\S\ref{sec:evaluation-metric}), the generation configurations and analysis (\S\ref{sec:configs}), and %
models (\S\ref{sec:models}).

\subsection{Data}
\label{sec:data}

Our study requires \textit{parallel} subjective judgments: multiple ratings of the same text, each linked to the demographic profile of the annotator who produced it. We therefore leverage the \textsc{DeMo} dataset \citep{orlikowski-etal-2025-beyond} which combines five annotator-level corpora and maps self-reported demographics onto four dimensions: \textit{gender}, \textit{age}, \textit{race}, and \textit{education}.
Since each annotation is aligned with the annotator’s profile, we can construct the gold label distribution of judgments for any text and demographic group. 

We retain three of the five tasks in \textsc{DeMo}, all rated on a five-point ordinal scale:\footnote{We exclude \emph{safety} (DICES-350; \citealp{aroyo2023dices}), which uses a three-way categorical scale and generational cohorts rather than age brackets, and \emph{sentiment} \citep{diaz2018addressing}, which recruited only annotators over $50$.}

\begin{itemize}[leftmargin=1em]
    \item \textbf{Intimacy}: rating how intimate a Twitter post is, from \textit{not intimate at all} to \textit{very intimate} (MINT; \citealp{pei-etal-2023-semeval});
    \item \textbf{Offensiveness}: rating how offensive a Reddit comment is, from \textit{not offensive at all} to \textit{very offensive} (POPQUORN; \citealp{pei-jurgens-2023-annotator});
    \item \textbf{Politeness}: rating how polite a workplace email is, from \textit{not polite at all} to \textit{very polite} (POPQUORN).
\end{itemize}

Table \ref{tab:example-distributions} shows an example of an offensiveness entry and the corresponding judgment distribution of two annotator groups (men and women).

\begin{table}[ht!]
\centering
\footnotesize
\begin{tabular}{@{}lcccccc@{}}
\toprule
\multicolumn{7}{@{}p{0.97\columnwidth}@{}}{
Text: \textit{``thats fucking hilarious, but also sad at the same time because I know it won't
change anything. Here's to hoping these laws get struck down, and old pseudo-religious trying to
win political brownie points men stop trying to tell women what they can and can't do with their
bodies''}} \\
\toprule
 & A & B & C & D & E & $s$ \\
\midrule
\textit{Humans} \\
Men ($n{=}4$) & $0$ & $2$ & $2$ & $0$ & $0$ & \\
Women ($n{=}4$) & $2$ & $0$ & $1$ & $0$ & $1$ & \\
\midrule
\textit{Llama 70B}\\
Unconditioned & $.00$ & $.07$ & $.40$ & $.52$ & $.01$ & --- \\
Man profile & $.12$ & $.46$ & $.28$ & $.13$ & $.01$ & $0.91$ \\
Woman profile & $.84$ & $.13$ & $.02$ & $.01$ & $.00$ & $0.67$ \\
\bottomrule
\end{tabular}
\caption{Offensiveness example. A--E are 5-point ratings from \textit{not offensive at all} to \textit{very offensive}; $s$ measures distance between rating distributions (see \ref{sec:evaluation-metric}). Human rows report annotator counts by gender ($n{=}4$ each). Llama~70B rows are predicted probabilities, with $s$ computed for man/woman profiles against the corresponding human distribution.}

\label{tab:example-distributions}
\end{table}

For our analysis, we remove categories too sparsely represented to support a stable per-group label distribution. 
The complete filtering procedure is described in Appendix~\ref{app:filtering}.
Statistics about the resulting dataset are reported in Table~\ref{tab:tasks}, whereas the demographic categories retained after filtering are reported in Table~\ref{tab:demographics}.

\begin{table}[ht!]
\setlength{\tabcolsep}{2.2pt}
\centering
\footnotesize
\begin{tabular}{llrrr}
\textbf{Task} & \textbf{Genre} & \textbf{Texts} & \textbf{Judgments} & \textbf{Annotators} \\
\toprule
Intimacy      & tweets          & 1,992  & 11,355  & 237   \\
Offensiveness & comments        & 1,500  & 12,489  & 251   \\
Politeness    & emails          & 3,718  & 23,896  & 483   \\
\midrule
\textbf{Total} &                & \textbf{7,210} & \textbf{43,402} & \textbf{971} \\
\bottomrule
\end{tabular}
\caption{Statistics of the 3 tasks retained from the \textsc{DeMo}
dataset, after filtering (see \S~\ref{sec:data}). The average number of judgements per entry is 6.02. }
\label{tab:tasks}
\end{table}

\begin{table}[ht!]
\setlength{\tabcolsep}{2pt}
\centering
\footnotesize
\begin{tabular}{ll}
\textbf{Dimension} & \textbf{Values} \\
\toprule

Gender (2) & Man, Woman \\
Age (8) & \begin{tabular}[t]{@{}l@{}}18--24, 25--29, 30--34, 35--39,\\40--44, 45--49, 50--59, 60--69\end{tabular} \\
Race (3) & \begin{tabular}[t]{@{}l@{}}Asian, Black, White\end{tabular} \\
Education (3) & \begin{tabular}[t]{@{}l@{}} High school or below, \\ College degree, Graduate degree
\end{tabular} \\
\bottomrule
\end{tabular}
\caption{Demographic dimensions and values used in our experiments.
Their cross-product ($2 \times 8 \times 3 \times 3$) defines the space of demographics we consider.
}
\label{tab:demographics}
\end{table}

\subsection{Evaluation Metrics}
\label{sec:evaluation-metric}

\paragraph{Distributions as targets.}

We evaluate a judge against the full distribution of human labels, at the level of demographic groups. Let $x_i$ be a text and $g$ a demographic profile, i.e. a combination of values for one or two of the four dimensions. 
The human reference $q^{(i,g)}$ is the distribution over the $K{=}5$ scale points of the labels that annotators matching $g$ assign to $x_i$. Each generation configuration yields a model distribution $p^{(i,g)}$ over the same points, from the log-probabilities the model assigns to the five answer tokens.
Evaluating a model thus consists in measuring, for each $(x_i, g)$ pair, how close $p^{(i,g)}$ is to $q^{(i,g)}$ (\S\ref{sec:configs-demographic}).

Table~\ref{tab:example-distributions} shows why we compare distributions rather than aggregated labels. 
Mapping the ordinal points A--E to equally spaced values 
in $[0,1]$ (i.e., $\{0, 0.25, 0.5, 0.75, 1\}$),
we can summarize a distribution by its \emph{mean rating}, its expected value under this mapping.
The two groups in the example have the same mean rating ($0.375$), so an averaged label would make the judge equally close to both. 
However, the distributions differ: men concentrate on the middle of the scale, women split between its two ends. Unlike the score $s$ defined below, the mean rating treats the scale as interval. We use it only to report the direction and magnitude of conditioning shifts.

\paragraph{Comparing ordinal distributions.}
Because our labels are ordinal, the discrepancy between two distributions should grow with the distance over which probability mass is misplaced. Divergence measures such as KL treat labels as nominal, penalizing misplaced mass equally regardless of where in the scale it lands.
We therefore use the Earth Mover's Distance (EMD) \citep{Rubner2000EMD}, the probability mass that must be moved to turn one distribution into the other, weighted by the ordinal distance moved:\footnote{EMD is used in NLP for subjective tasks, from document similarity \citep{kusner2015word} to aligning model and human response distributions on ordinal survey scales \citep{santurkar2023whose, tjuatja-etal-2024-llms}.}

    $$\mathrm{EMD}(p, q) = \sum_{k=1}^{K-1} \Bigl|\, \textstyle\sum_{j \leq k} p_j - \sum_{j \leq k} q_j \,\Bigr|.$$

Following \citet{santurkar2023whose}, we report it as a bounded similarity score, inverted so that higher is better:

$$s(p, q) = 1 - \frac{\mathrm{EMD}(p,q)}{K-1}.$$

Because the maximum EMD on a $K$-point scale is $K-1$, reached when the two distributions place all their mass on opposite extremes, $s$ ranges from $0$ (maximal divergence) to $1$ (perfect match) and is comparable across tasks and configurations. 
To isolate the effect of demographic conditioning, we score both the unconditioned and the conditioned prediction (see \S\ref{sec:configs}) for the same text against the same group reference:

    $$\Delta s = s\bigl(p^{(i,g)}, q^{(i,g)}\bigr) - s\bigl(p^{(i)}_{\emptyset}, q^{(i,g)}\bigr).$$

where $p^{(i)}_{\emptyset}$ is the model's prediction for $x_i$ without a profile. While $s$ measures how closely a judge matches a group, $\Delta s$ isolates the effect of the profile alone: $\Delta s > 0$ means conditioning moved the judge toward the group, $\Delta s < 0$ that it moved the judge away from the group it was instructed to represent.

\paragraph{Aggregation and uncertainty.}

Both $s$ and $\Delta s$ can be affected by the shape of the prediction and reference distributions: a more spread-out prediction tends to score better against a more spread-out reference, even without better group alignment.
This can distort comparisons in two ways:
(i) on the human labels side, groups can have different numbers of annotators per item, so some reference distributions are more sparse than others; (ii) on the model side,
instruction-tuning is known to sharpen answer-token distributions \citep{santurkar2023whose, durmus2023measuring, sorensen-etl-2024-pluralistic-alignment}, so any comparison of \textit{base} and \textit{instruct} model variants risks being an artifact of sharpness. We therefore validate our results against mode accuracy, which is unaffected by distribution sharpness.
We average scores within each cell (a model, configuration, task, and demographic group) and then macro-average cell means with equal weight, preventing groups with more observations from dominating the average. Confidence intervals are $95\%$ percentile bootstraps over cells (2,000 resamples).
We validate all group comparisons against two independent spread-insensitive controls: density-matched references and mode accuracy. Mode accuracy also mitigates the concern that first-token probabilities may diverge from the model's generated answer \citep{wang-etal-2024-answer-c}. We verify token coverage but do not compare against generated text (Appendix \ref{app:robustness-checks}). 

\subsection{Experimental Design}
\label{sec:configs}

We elicit a judgment from a model in a single forward pass and read off the full distribution it places on the five scale points, rather than sampling a discrete answer. We follow the prompt format of \citet{orlikowski-etal-2025-beyond}: each prompt presents the task question, the text to be judged, and the five labeled options \mbox{(A)--(E)} in order, and ends with an assistant-turn prefix constraining the model's next token to be one of the option letters (Figure~\ref{fig:methodology}).\footnote{The order is fixed, but the models show no choice-position bias \citep{zheng2024large}: pooled over the 23 models, option mass tracks the human labels ($r = 0.83$) rather than letter order, and the modal option \mbox{(A)} $(0.275$ against $0.212$ for \mbox{(B)})\ is also the modal human label $(0.323)$.}
All results in \S\ref{sec:results-groups} use this single template, since conditioned and unconditioned predictions must be scored under the same wording to be paired. 
Because prompt format and wording could 
play a relevant role in the distributions \citep{beck-etal-2024-sensitivity}, we 
also run all $23$ models under a structurally different template, an interview-style profile in which the annotator states their own demographics \citep{lutz-etal-2025-prompt} (see Appendix~\ref{app:prompts}). 

Following prior work that reads judgment distributions directly from answer-token probabilities \citep{santurkar2023whose, durmus2023measuring}, we take the predicted distribution $p$ to be the softmax over the log-probabilities of the five option tokens \texttt{A}--\texttt{E}.\footnote{We retrieve the top-20 next-token log-probabilities and read each option's first token, assigning $-\infty$ to any option absent from that set. } 
The configurations below differ only in \textit{(i)} the demographic profile, if any, prepended to the prompt, and \textit{(ii)} the human reference against which each prediction is scored (\S\ref{sec:evaluation-metric}).

\paragraph{Unconditioned Baseline.}
\label{sec:configs-baseline}

In this configuration the prompt carries no demographic information (Figure~\ref{fig:methodology}, without the profile line), thus the model judges without being assigned a perspective.
For each text, the baseline prediction is compared with the human label distribution of every demographic group that annotated it. %
This unconditioned setting %
(i) provides the reference point for score difference  $\Delta s$ (\S\ref{sec:evaluation-metric}), isolating the effect of adding a demographic profile to the prompt, and %
(ii) reveals possible models' \emph{default} demographic alignment (i.e., the perspective it tends to adopt in the absence of demographic conditioning). %

\paragraph{Demographic Conditioning.}
\label{sec:configs-demographic}

We prompt 
the model to judge the text from the perspective of a specific demographic group, i.e.\ a \textit{profile} (Figure~\ref{fig:methodology}), and score its prediction against that group's human label distribution.
A profile assigns a value to one or two of the four dimensions of \S\ref{sec:data}: gender, age, race, and education. We consider the four \emph{single-attribute} configurations and the six \emph{two-attribute} combinations.\footnote{We stop at two attributes, since narrower profiles leave too few annotators per cell to estimate a stable reference.} The former isolate each dimension's contribution, the latter test whether intersectional profiles help or hurt. Across the ten configurations this yields 
268K $(x_i, g)$ pairs per model. For each, we compute $s$ twice against the same reference $q^{(i,g)}$ (once for the conditioned prediction $p^{(i,g)}$, once for the unconditioned $p^{(i)}_{\emptyset}$) and take their difference as $\Delta s$. 

\begin{figure*}[t]
\centering
\includegraphics[width=\textwidth]{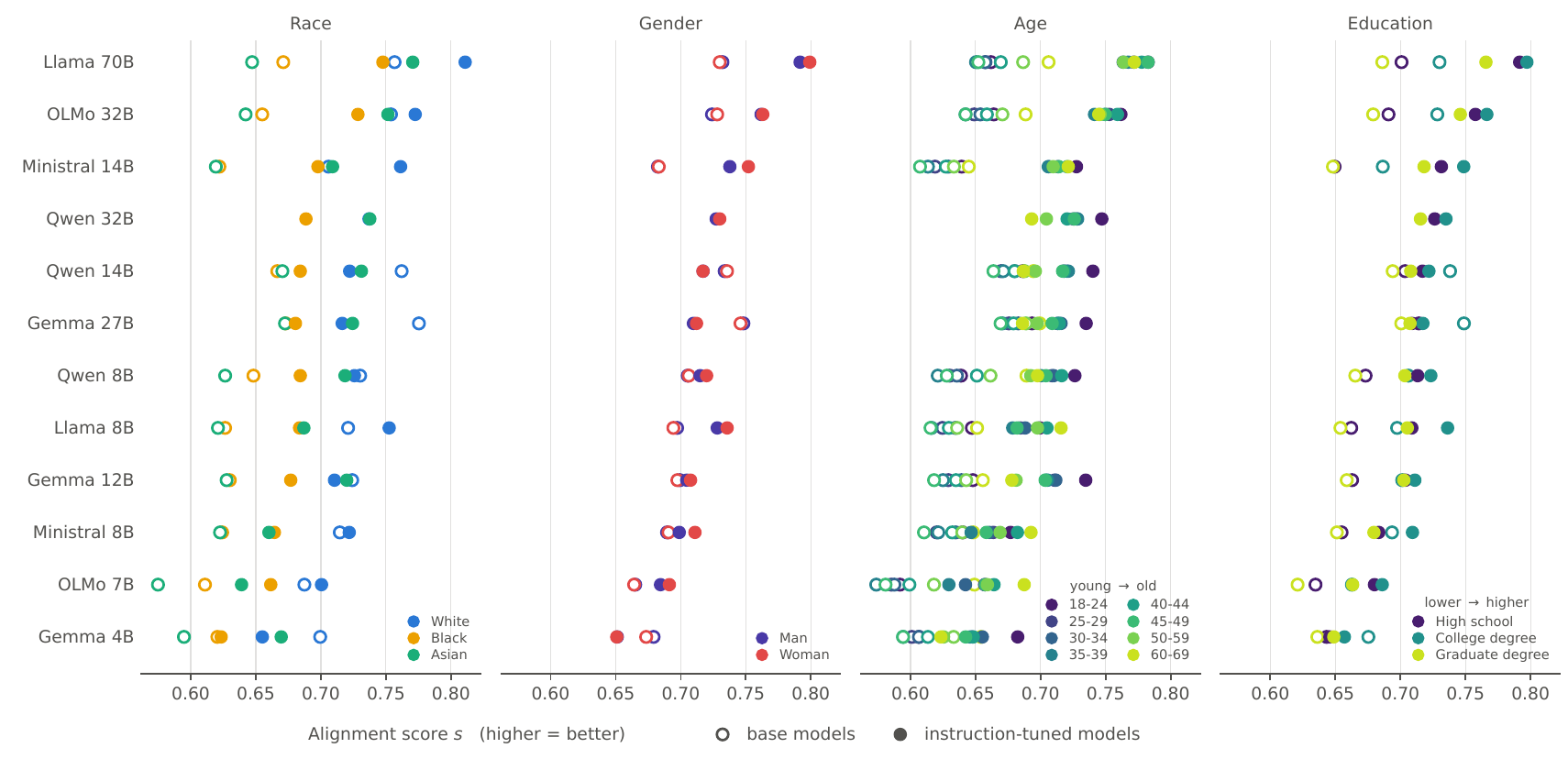}
\caption{Alignment of the unconditioned baseline prediction with each demographic group's human label distribution, one panel per dimension and one row per model. 
Hollow markers are \textit{base} models, filled markers \textit{instruct} models. Color identifies the group. 
}
\label{fig:group-s}
\end{figure*}

\subsection{Models}
\label{sec:models}

We evaluate models from five families, ranging from 4B to 70B parameters: Gemma~3 4B, 12B, and 27B \citep{gemmateam2025gemma3technicalreport}; Llama~3.1 8B and 70B \citep{grattafiori2024llama3herdmodels}; Ministral~3 8B and 14B \citep{liu2026ministral3}; OLMo~3 7B and 32B \citep{olmo2026olmo3}; and Qwen~3 8B, 14B, and 32B \citep{yang2025qwen3technicalreport}. We evaluate models in both \textit{base} and instruction-tuned form,\footnote{Instruction-tuned models receive the prompt in their chat template, with the final assistant turn left open at the answer prefix. \textit{Base} models receive the same messages as plain concatenation, so each is prompted in its native format.} except Qwen~3 32B, which is available only instruction-tuned, for 23 models in total. Variation in size within each family lets us study scale while keeping the model family fixed. The 11 matched \textit{base}--\textit{instruct} pairs isolate the effect of 
instruction-tuning.

\section{Results and Discussion}
\label{sec:results-discussion}
We organize the results around the three research questions of \S\ref{sec:intro}.%

\subsection{RQ1: Default Profiles of LLM-Judges}
\label{sec:results-default}

In line with \citet{schafer-etal-2025-demographics}, who found judges furthest from Black annotators and unaffected by gender on two proprietary models, we find a shared default profile in open-weight models' label distributions.

\paragraph{All models align best with \textit{White} and \textit{college-educated} annotators, equally well with \textit{men} and \textit{women}, and differ from one another in how they relate to age.}
Figure~\ref{fig:group-s} reports the alignment of the unconditioned predictions with each demographic group's label distributions, one panel per dimension.
Race shows the largest gaps: 
all base models and the majority of \textit{instruct} models match White annotators more closely than Black and Asian ones,
with an average White--Black gap of $0.089$ for base models and $0.047$ for \textit{instruct} models. 

For base judges, this corresponds to moving more than one third of the probability mass by one response step away from Black annotators' judgments.\footnote{The ordering is robust to annotator-count differences for instruction-tuned models, and holds for most base models under matching, even as the gap sizes shrink (Appendix~\ref{app:density-matching}). Such group-level differences are visible only under a distributional comparison, while scoring against a single aggregate label would collapse (\S\ref{sec:evaluation-metric}).}
Education is the most consistent dimension: pooled across tasks, every model is closest to college-educated annotators, ahead of annotators with a high-school education.
Gender, by contrast, shows no default: the markers largely overlap, with mean absolute gaps below $0.01$.
Age is the only dimension on which models disagree, and the split tracks model type: base models are generally closest to the oldest group, whereas most \textit{instruct} models are closest to the youngest.
This is a first indication that %
instruction-tuning changes whose perspective a judge adopts even before any profile is assigned. We investigate the impact of %
instruction-tuning further in \S\ref{sec:results-conditioning}. These defaults are largely stable across tasks: race and gender patterns hold on all three, while age (for all models) and education (for \textit{instruct} models only) vary more by task (see Appendix \ref{app:default-groups}).
Appendix~\ref{app:robustness-checks} verifies that these orderings are not artifacts of distributional spread.

\paragraph{An unconditioned judge is not a neutral annotator.}

LLMs are often used as substitutes for human annotators, but an unconditioned model should not be treated as an average annotator. 
Without any demographic profile, every instruction-tuned judge aligns more closely with White than Black or Asian annotators, and with college-educated than less-educated annotators. %
Thus, using an unconditioned 
judge as a generic annotator introduces a systematic demographic perspective. This confirms, on open-weight judges and distributional scoring, the conclusion of \citet{schafer-etal-2025-demographics} that LLMs do not represent all social groups equally.
The analyses that follow ask what happens 
when a profile is assigned: whether conditioning corrects it or compounds it.

\subsection{RQ2: The Effect of Demographic Conditioning}
\label{sec:results-conditioning}

\paragraph{Demographic conditioning helps base models, but not \textit{instruct} models (on average).}

\begin{figure}[ht!]
\centering
\includegraphics[width=\columnwidth]{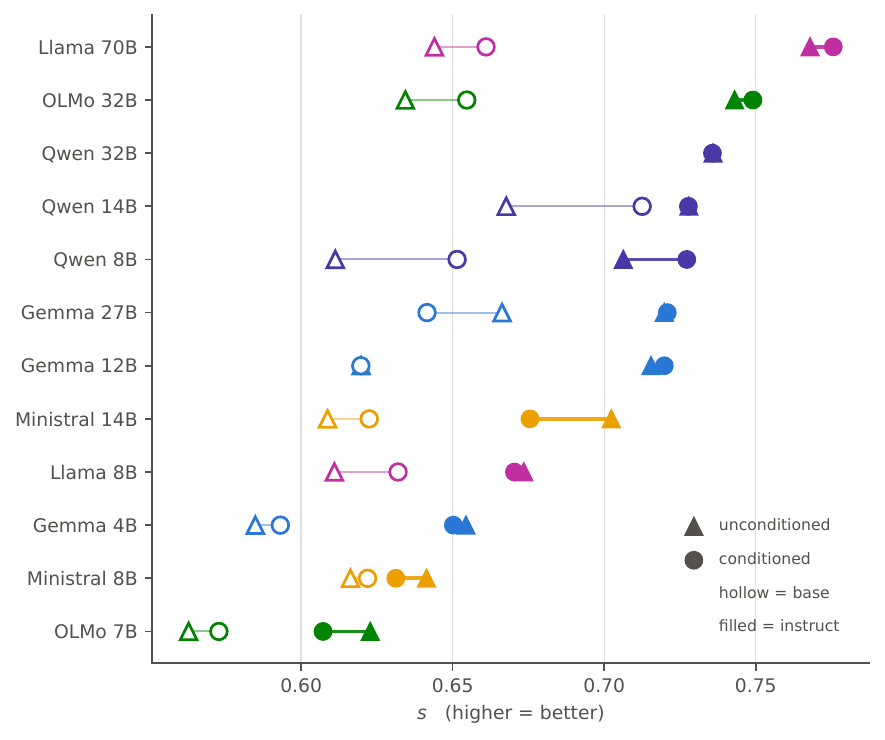}
\caption{Alignment with annotator groups, unconditioned (triangle) vs. conditioned (circle), for the 
matched base models (hollow) and the
instruction-tuned judges (filled). 
The
horizontal distance between a row's triangle and circle represents the effect of conditioning for that model ($\Delta s$).}
\label{fig:uncond-vs-cond}
\end{figure}

Figure~\ref{fig:uncond-vs-cond} compares the conditioned and unconditioned predictions of each judge against the same group references.
Assigning a profile improves 10 of 11 base models, while among the 12 \textit{instruct} models it helps five, leaves two unchanged, and hurts the rest: on average, conditioning appears to have %
little to no effect on instruction-tuned models.

\paragraph{The base model gain is not an artifact of spread-out predictions.}

Base models distribute their probability more evenly across the five options than \textit{instruct} models, which tend to concentrate it on one or two.\footnote{Measured as entropy normalized by its maximum $\log K$, so that $0$ is a one-point distribution and $1$ a uniform one: $0.919$ on average for base models, \ $0.274$ for the \textit{instruct}.} 
This asymmetry admits an alternative explanation for the base model gain: a spread-out prediction necessarily overlaps part of any reference distribution, so conditioning could improve base scores simply by reshaping an almost-flat prediction, without moving the judge toward the group's actual judgments.
We rule out this explanation using mode accuracy, which ignores probability spread and gives credit only when the model's most likely option matches the group's majority label. Under this metric, the mean conditioning gain for base models increases from $+0.014$ to $+0.032$ and remains positive for $10$ of $11$ models (see Appendix~\ref{app:top1}). \textit{Instruct} models remain near zero under both metrics.

\paragraph{Neither scale nor instruction-tuning makes judges more steerable.}

Larger models align better with annotators: within the instruction-tuned pool, $r = 0.85$ between log parameter count and $s$, and \textit{instruct} models start $0.077$ ahead of the \textit{base} ones in the unconditioned setting. 
Yet neither gain carries over to conditioning: log parameter count and $\Delta s$ are essentially uncorrelated across the 23 judges ($r = 0.016$), and the largest gains fall on the \textit{base} Qwen models regardless of scale (Qwen 8B $+0.041$, Qwen 14B $+0.044$).

Better judges, however, are not \textit{more steerable} judges. A judge can be wrong about a text in two ways: it can misjudge the text itself, which puts it off for every group at once, or it can miss the way one group departs from the others. %
Instruction-tuning only improves the first, and every group's score rises accordingly.

\paragraph{The \textit{instruct} models' null effect is not uniform across tasks.}

For base models, conditioning helps on all three tasks.
For \textit{instruct} models, the $\Delta s$ is below $0.01$ on politeness and intimacy but turns \emph{negative} on offensiveness ($-0.011$ on average, down to $-0.080$ for Ministral 14B). Appendix~\ref{app:cond-tables} reports the per-model effects: on the one task where demographic perspective shows the greatest variability, assigning a profile makes several \textit{instruct} judges \emph{worse} at representing the group they are instructed to represent.

However, the almost flat average conceals relevant per-group behavior, both positive and negative \citep{simpson1951interpretation,blyth1972simpson}. 
The next section decomposes it, asking which groups conditioning helps and which it hurts.

\begin{figure*}[ht!]
\centering
\includegraphics[width=\textwidth]{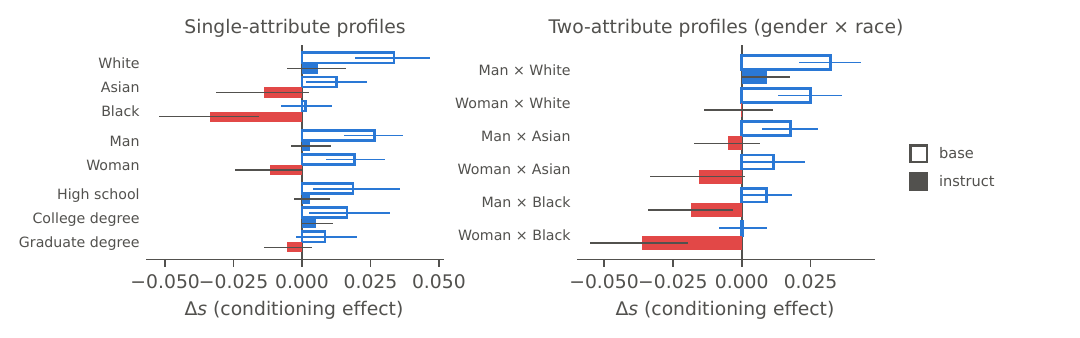}
\caption{Conditioning effect ($\Delta s$) by demographic group
for the 11 models released in both configurations. Bars are hollow for \textit{base}, filled for \textit{instruct}, colored by sign. Whiskers are $95\%$ cluster-bootstrap intervals over judges. Left: single attributes. Right: gender$\times$race. Age is omitted, as no bracket differs from zero for \textit{instruct} models.}
\label{fig:delta-by-group}
\end{figure*}

\begin{figure*}[ht!]
\centering
\includegraphics[width=\textwidth]{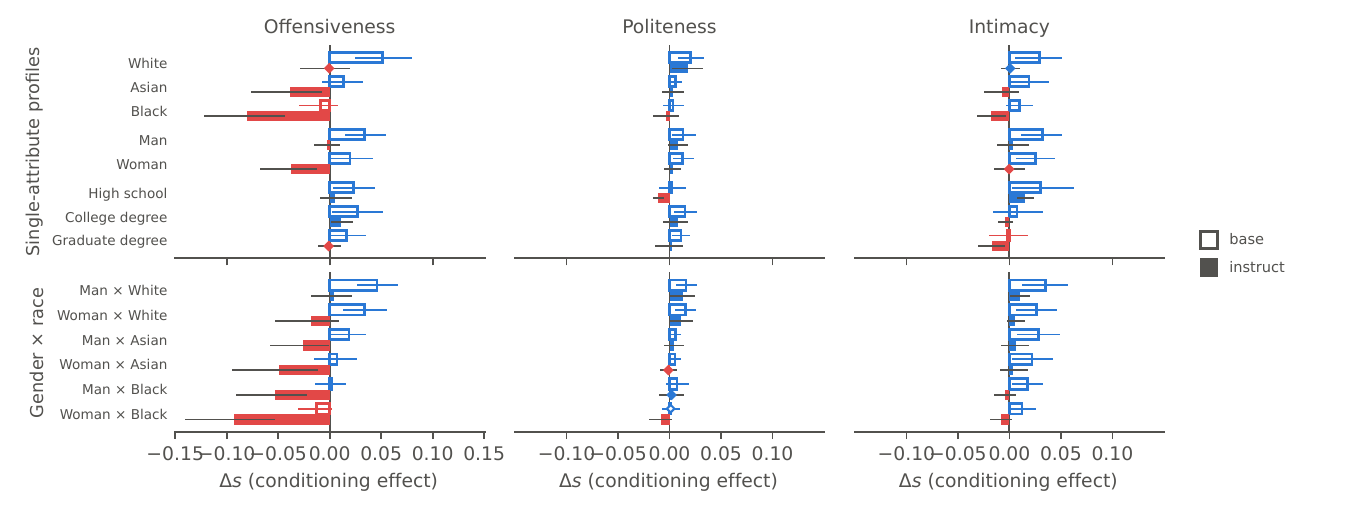}
\caption{As Figure~\ref{fig:delta-by-group}, one column per task. Top row: single attributes. Bottom row: gender$\times$race. Diamonds mark effects below $0.002$, too small to draw.}
\label{fig:delta-by-group-bytask}
\end{figure*}

\subsection{RQ3: Conditioning Helps Majority Groups and Hurts Minority Groups}

\label{sec:results-groups}

\paragraph{Instruction-tuning makes demographic conditioning uneven across groups.}
Figure~\ref{fig:delta-by-group} shows the conditioning effect separately for each demographic group. 
For \textit{base} models, conditioning is positive for every group: 
assigning a profile moves the judge toward that group's judgments regardless of which group it is.
With instruction-tuned models, the effect splits:
conditioning remains positive for White, Man, College degree, and High school, but becomes negative for Asian, Woman, Graduate degree, and Black.  The largest harm falls on Black annotators ($-0.033$): instructing the judge to answer as a Black annotator makes it \emph{less} aligned with Black annotators' judgments than giving it no profile at all.

Intersectional profiles do not combine additively: the joint effect is attenuated toward the gender marginal, landing above the sum of the two single-attribute effects. Man\,$\times$\,White benefits ($+0.009$), while against Black, the near-zero Man effect dilutes the harm (Man\,$\times$\,Black $-0.019$) while Woman deepens it (Woman\,$\times$\,Black $-0.036$).

\paragraph{The harm to minority profiles is concentrated on offensiveness.}

Figure~\ref{fig:delta-by-group-bytask} splits the same breakdown by task. Unlike the defaults of \S\ref{sec:results-default} and the pooled effects of \S\ref{sec:results-conditioning}, which apply similarly to all three tasks, the harm to minority groups is more evident in offensiveness.
On offensiveness, conditioning an \textit{instruct} judge reduces alignment with Black annotators by $-0.074$, with Asian annotators and with women by $-0.035$. 
Every gender\,$\times$\,race profile containing a minority attribute is negative, with \emph{Woman\,$\times$\,Black} at $-0.087$, the largest effect in the study.
The pattern is weaker on intimacy, where \emph{Black} is the only group with a reliable negative effect ($-0.018$), and absent on politeness, where no minority profile is harmed.
\textit{Base} models show none of this task specificity: their conditioning effect is positive for nearly every group on all three tasks.
Thus, the asymmetry is not a general property of all LLMs under demographic conditioning. It is rather introduced by %
instruction-tuning and expressed on the task where demographic perspective is most contested.

\paragraph{Prompting moves the judge toward a stereotype, not toward the group.}

\begin{figure*}[ht]
\centering
\includegraphics[width=\textwidth]{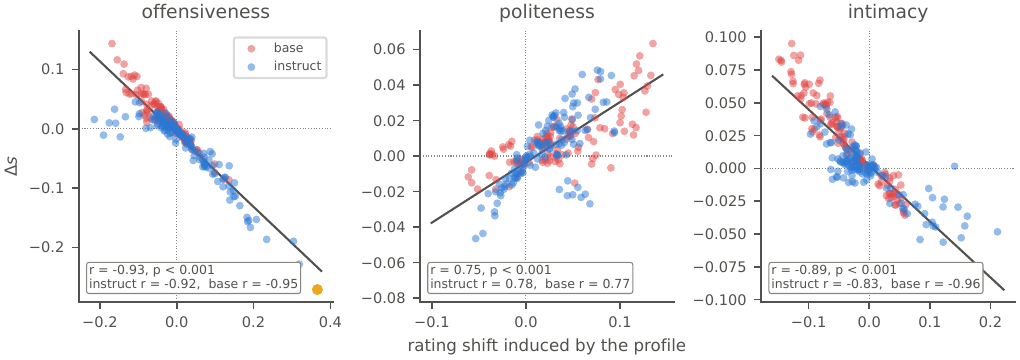}
\caption{One point per judge and profile, one panel per task. The horizontal axis shows the change in predicted rating; the vertical axis shows the change in group alignment ($\Delta s$). The yellow point marks Ministral 14B with \emph{Woman$\times$Black}. Annotations report Pearson $r$, overall and by model pool.}
\label{fig:severity-mechanism}
\end{figure*}

\begin{table}[t]
\centering
\footnotesize
\setlength{\tabcolsep}{5pt}
\begin{tabular}{@{}l c c c@{}}
\textbf{Group} & \textbf{Human} & $\Delta$ \textbf{base} & $\Delta$ \textbf{instruct} \\
\toprule
\multicolumn{4}{@{}l}{\textit{Offensiveness}}\\
White & $-0.018$ & $-0.070$ & $-0.026$ \\
Black & $+0.052$ & $+0.005$ & $+0.080$ \\
Asian & $-0.003$ & $-0.031$ & $+0.010$ \\
Man   & $+0.006$ & $-0.044$ & $-0.004$ \\
Woman & $+0.003$ & $-0.026$ & $+0.039$ \\
\addlinespace
\multicolumn{4}{@{}l}{\textit{Politeness}}\\
White & $-0.001$ & $+0.054$ & $+0.051$ \\
Black & $+0.028$ & $+0.022$ & $+0.003$ \\
Asian & $-0.027$ & $+0.047$ & $+0.029$ \\
Man   & $+0.005$ & $+0.027$ & $+0.020$ \\
Woman & $+0.001$ & $+0.029$ & $+0.009$ \\
\addlinespace
\multicolumn{4}{@{}l}{\textit{Intimacy}}\\
White & $+0.005$ & $-0.044$ & $-0.017$ \\
Black & $+0.048$ & $-0.025$ & $+0.030$ \\
Asian & $-0.025$ & $-0.036$ & $+0.001$ \\
Man   & $-0.018$ & $-0.047$ & $-0.005$ \\
Woman & $+0.011$ & $-0.039$ & $+0.009$ \\
\bottomrule
\end{tabular}
\caption{Group deviation from the pooled human mean and the rating shift induced by conditioning for all models ($\Delta$), by task and single-attribute group (mean rating on the [0,1] scale described in \S\ref{sec:evaluation-metric}). Faithful conditioning would give $\Delta~\approx$~Human. 
}
\label{tab:shift-vs-dev}
\end{table}

Demographic conditioning is effective when it moves the judge's rating by the amount the group actually differs from the mean of all annotators.
The two shifts are reported in Table~\ref{tab:shift-vs-dev}. %
Two patterns emerge. On offensiveness, Black annotators rate $+0.052$ above the mean, yet the \emph{Black} profile moves \textit{instruct} judges by $+0.080$, about $50\%$ more than the real difference: conditioning amplifies the gap rather than reproducing it.
On politeness, \emph{Asian} is the only group rating below the mean ($-0.027$), yet its profile shifts judges upward 
($+0.029$ \textit{instruct}, $+0.047$ \textit{base}), opposite to the group it names.
As the table shows, the shift is largely insensitive to the group's actual deviation: on offensiveness it is upward for \textit{instruct} judges regardless of the group, and only \textit{base} judges, which overestimate offensiveness unconditioned, move back toward the human ratings.
In general, the average shift is upward, whether the group differs from the mean or not: the profile triggers the expectation that the named group perceives more offensiveness, not that group's actual judgments.
This confirms the stereotype activation reported for persona-prompted generation \citep{gupta2024bias, deshpande-etal-2023-toxicity}, and quantifies it, as against real annotator distributions, the shift can be measured against the group's actual deviation.
Intersectional profiles show the same pattern as the corresponding single-attribute ones. We report them in Appendix~\ref{app:faithfulness-intersect}.

\paragraph{Gender shows that judges shift even when there is no group-related difference.}
As shown in Table~\ref{tab:shift-vs-dev}, gender is the one dimension with no default (\S\ref{sec:results-default}) and no group deviation to reproduce: women rate at the pooled mean (e.g., $+0.003$ in offensiveness). A faithful judge conditioned on \textit{Woman} would therefore not move. Instead the profile shifts instruction-tuned judges upward (e.g., $+0.039$ in offensiveness).
Here, on the contrary, the distributions unduly move because of the models' stereotyped expectations of the group.

\paragraph{Conditioning shifts ratings, not perspectives.}

Figure~\ref{fig:severity-mechanism} shows the correlation between shifts in the label distributions and $\Delta s$.
On offensiveness, the more conditioning raises a judge's rating, the more alignment it loses ($r=-0.93$, $p<.001$). 
The same trend holds when comparing judges given the same profile (within-profile $r=-0.92$). 

\textit{Base} models tend to move downward, toward the human ratings, and therefore improve; \textit{instruct} models tend to move upward, away from them, and therefore worsen. Intimacy shows the same pattern.
Politeness reverses the direction. Judges initially underrate politeness, predicting $0.38$ on average against a human mean of $0.58$. Conditioning tends to push their ratings upward, so larger shifts improve alignment ($r=+0.75$). 
Yet this does not mean the judges recover each group's perspective: only 5 of the 11 profiles shift the judge in the same direction as the group's actual deviation from the overall human mean.
The gain therefore comes mainly from correcting the judges' general underestimation of politeness, not from reproducing group-specific judgments.

Across all three tasks, the mechanism is the same: demographic conditioning pushes \textit{all} predictions in a direction that is not reliably tied to the group being represented. It helps when that push happens to move the judge toward the human ratings, and hurts when it moves the judge away.

\section{Conclusions}
\label{sec:conclusions}

Across 23 open-weight LLM judges, we asked whether conditioning a judge on an annotator's demographic profile brings its judgments closer to that group's. %
An unconditioned judge is not perspective-neutral: every model aligns more closely with White than with Black or Asian annotators, and with college-educated annotators than with either other band. Conditioning an instruction-tuned judge appears to have no effect on average. However, that average hides gains for the groups the judge already leans toward, and losses for minority groups. The harm concentrates on offensiveness, is sharpened rather than repaired by intersectional profiling, and is not recovered by %
model size. Comparing each instruction-tuned judges with their base counterparts locates the asymmetry in %
instruction-tuning: conditioning positively impacts \textit{base} models %
for every group, whereas with instruction-tuned models the profile does not activate the group's perspective but an expectation about it, exaggerating differences where they exist and introducing them where they do not.
Thus the practical question is not simply \textit{whether} demographic conditioning helps, but \textit{whom} it helps: its benefits are smallest for the very groups such interventions are meant to represent.

\section*{Limitations}
\label{sec:limitations}

Our findings rest on three English tasks from one annotator-level corpus \citep{orlikowski-etal-2025-beyond}, so we cannot claim the pattern transfers to other languages or annotation schemes.
Those corpora are public, and any leakage into pretraining would favor the alignment we measure. Per-group distributions thin as profiles narrow, so we condition on at most two attributes, and minority groups carry fewer parallel annotations \citep{difallah2018demographics, prabhakaran2021releasing}. 

\section*{Ethics Statement}

No new annotation was collected to conduct the analysis described in this work and no  annotator is identifiable.
We use coarse demographic categories only to define the group reference distributions, not as proxies for individual perspectives.
Our results caution against that stronger reading, since conditioning helped the groups judges already favour while harming the rest.
Thus, demographic prompting should not be treated as a substitute for recruiting annotators from the groups of interest. 



\bibliography{custom}

\bibliographystyle{acl_natbib}


%

\appendix

\section{Filtering of Under-Represented Categories}
\label{app:filtering}

As described in \S\ref{sec:data}, we remove annotations from demographic categories too sparsely represented in the source data to support reliable group-level comparison. Within each dimension we drop values contributed by too few annotators to estimate a stable per-group label distribution. 
Counting distinct annotators on the three retained tasks, this removes \emph{Non-binary} from gender, 
\emph{Hispanic/Latino} from race, 
\emph{Less than high school} from education, 
and the two oldest age brackets, $70$--$79$ and $80$--$89$, each 
populated only
by a single annotator. 

A cell is dropped if any value in its attribute conjunction is excluded, since the cell's human reference is defined by the full conjunction. Filtering is applied before any scoring, so unconditioned and conditioned predictions are evaluated against references drawn only from retained categories.

\section{Prompt Templates}
\label{app:prompts}

Every task uses the same two-message template: the user message states the task, presents the item, and lists the five options in ascending order, and the assistant message is a continued prefix, so the next token is the option letter. A conditioned prompt prepends the \texttt{**Your Profile**} line, one clause per active attribute in the fixed order race, age, gender, education.

\begin{center}
\begin{minipage}{\columnwidth}
\begin{mdframed}[backgroundcolor=dialoguebg, linecolor=dialogueframe, linewidth=1pt, roundcorner=5pt, innerleftmargin=2pt, innerrightmargin=2pt, innertopmargin=2pt, innerbottommargin=2pt, nobreak=true]
\footnotesize\raggedright
\texttt{[user]}\\[0.3em]
\textbf{Your Profile}: In terms of race or ethnicity, you are White. You are 35 to 39 years old. 
\\[0.4em]
\textbf{Question}: Consider you read this comment on Reddit, how offensive do you think it is?\\[0.2em]
\textbf{Comment:}: \textit{<text>}\\[0.4em]
(A) not offensive at all\\
{[...]}\\
(E) very offensive\\[0.6em]
\texttt{[assistant]}\\[0.3em]
\textbf{Answer}: (
\end{mdframed}
\captionof{figure}{The prompt template (offensiveness, two-attribute profile). The final assistant turn is left open at the answer prefix (\S\ref{sec:configs}).}
\label{fig:prompt-template}
\end{minipage}
\vspace{8pt}   
\end{center}



The unconditioned prompt is the same with the \texttt{Your Profile} line removed, so both conditions put the same question to the model and differ only in whether a profile precedes it. For base models the same messages are concatenated in plain text, without chat-template.

The other two framings ask how intimate a Twitter \texttt{**Post**} is and how polite an \texttt{**Email:**} from a colleague is, with options from \emph{not intimate/polite at all} to \emph{very intimate/polite}. All eleven configurations are released with the code as fully rendered prompts.

\paragraph{Robustness to the template.}
To test whether the demographic conditioning effects depend on the wording, we repeat the experiment with an interview template in which the model states the profile itself \citep{lutz-etal-2025-prompt}. Question, items and options are unchanged; only the lines before them differ.



\begin{center}
\begin{minipage}{\columnwidth}
\begin{mdframed}[backgroundcolor=dialoguebg, linecolor=dialogueframe, linewidth=1pt, roundcorner=5pt, innerleftmargin=2pt, innerrightmargin=2pt, innertopmargin=2pt, innerbottommargin=2pt, nobreak=true]
\footnotesize\raggedright
\texttt{[user]}\\[0.3em]
\textbf{Interviewer}: Before we begin, tell me a little about yourself.\\[0.2em]
\textbf{You}: In terms of race or ethnicity, I am White.\\[0.2em]
\textbf{Interviewer}: Thank you. Now please answer the following.\\[0.4em]
\textbf{Question}: [...]\\[0.2em]
\textbf{Comment}: \textit{<text>}\\[0.4em]
(A) not offensive at all\\
{[...]}\\
(E) very offensive\\[0.6em]
\texttt{[assistant]}\\[0.3em]
\textbf{Answer}: (
\end{mdframed}
\captionof{figure}{The interview-style template: the annotator states the demographic attribute in first person instead of receiving it as an assigned profile.}
\label{fig:prompt-template-interview}
\end{minipage}
\vspace{8pt}   
\end{center}

\begin{figure}[h]
\centering
\includegraphics[width=\columnwidth]{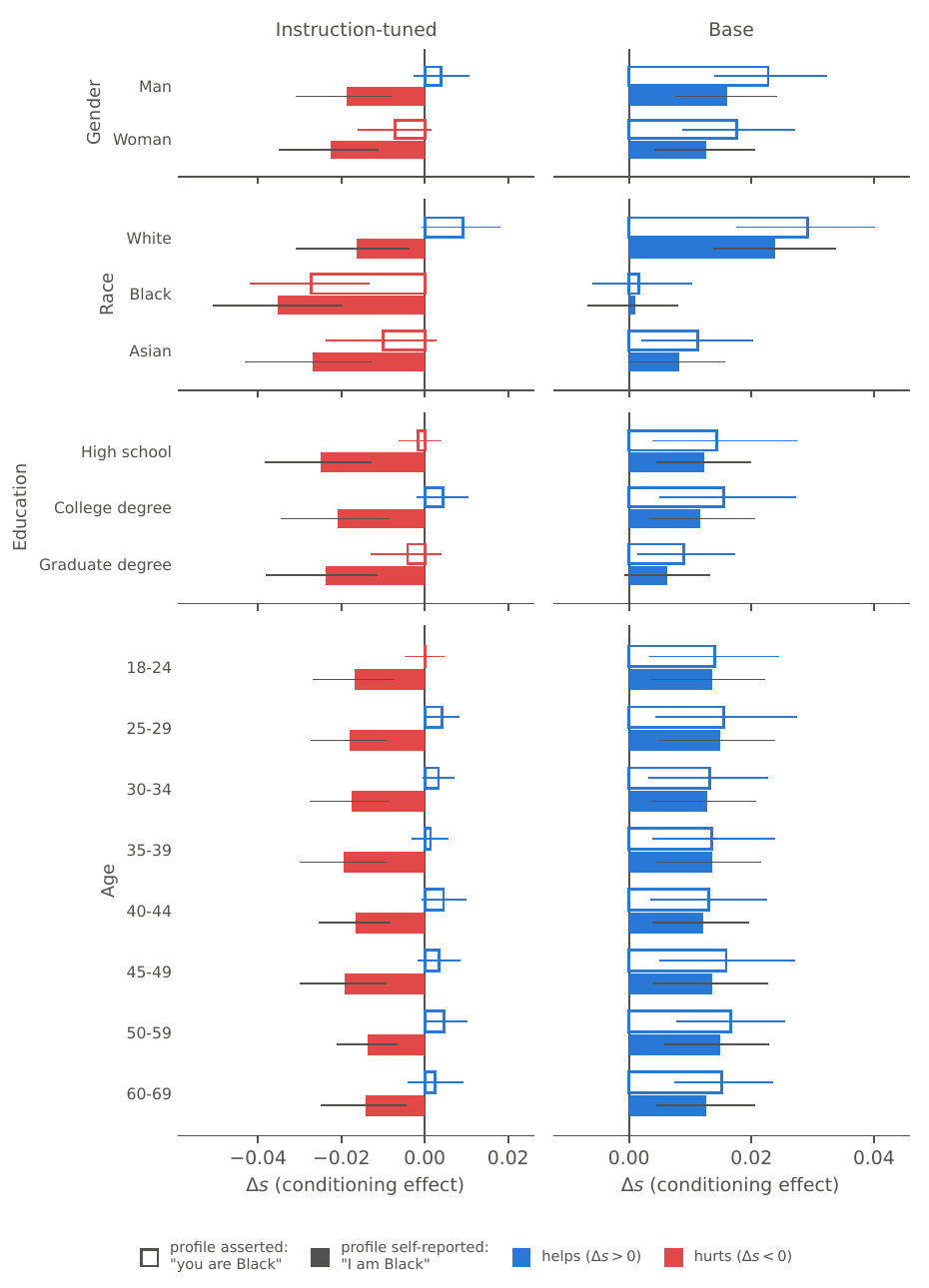}
\caption{The conditioning effect per group when the profile is asserted to the model against when the model states it itself. Bars are $\Delta s$, with $95\%$ cluster bootstrap intervals over judges.}
\label{fig:template-robustness}
\end{figure}

\noindent Its unconditioned arm drops the exchange and opens with \texttt{Interviewer: Please answer the following.}, so what separates the two arms is the profile alone.
Figure~\ref{fig:template-robustness} compares the conditioning effects across groups under the two templates.


The negative effects do not disappear under the alternative template. For \textit{instruct} models, conditioning becomes more negative across groups: for Black annotators, for example, $\Delta s$ changes from $-0.027$ to $-0.035$. On average, the self-report template lowers $\Delta s$ by $0.020$, making the effect negative for all sixteen groups. \textit{Base} models are much less sensitive to the change in wording: no group changes by more than $0.007$, and conditioning remains positive for all sixteen groups. 
The negative effects are therefore not specific to the wording of the main prompt, and are stronger when the model states the identity itself.

\section{Per-Group Alignment of the Unconditioned Judge}
\label{app:default-groups}

\begin{figure*}[th!]
\centering
\includegraphics[width=0.7\textwidth]{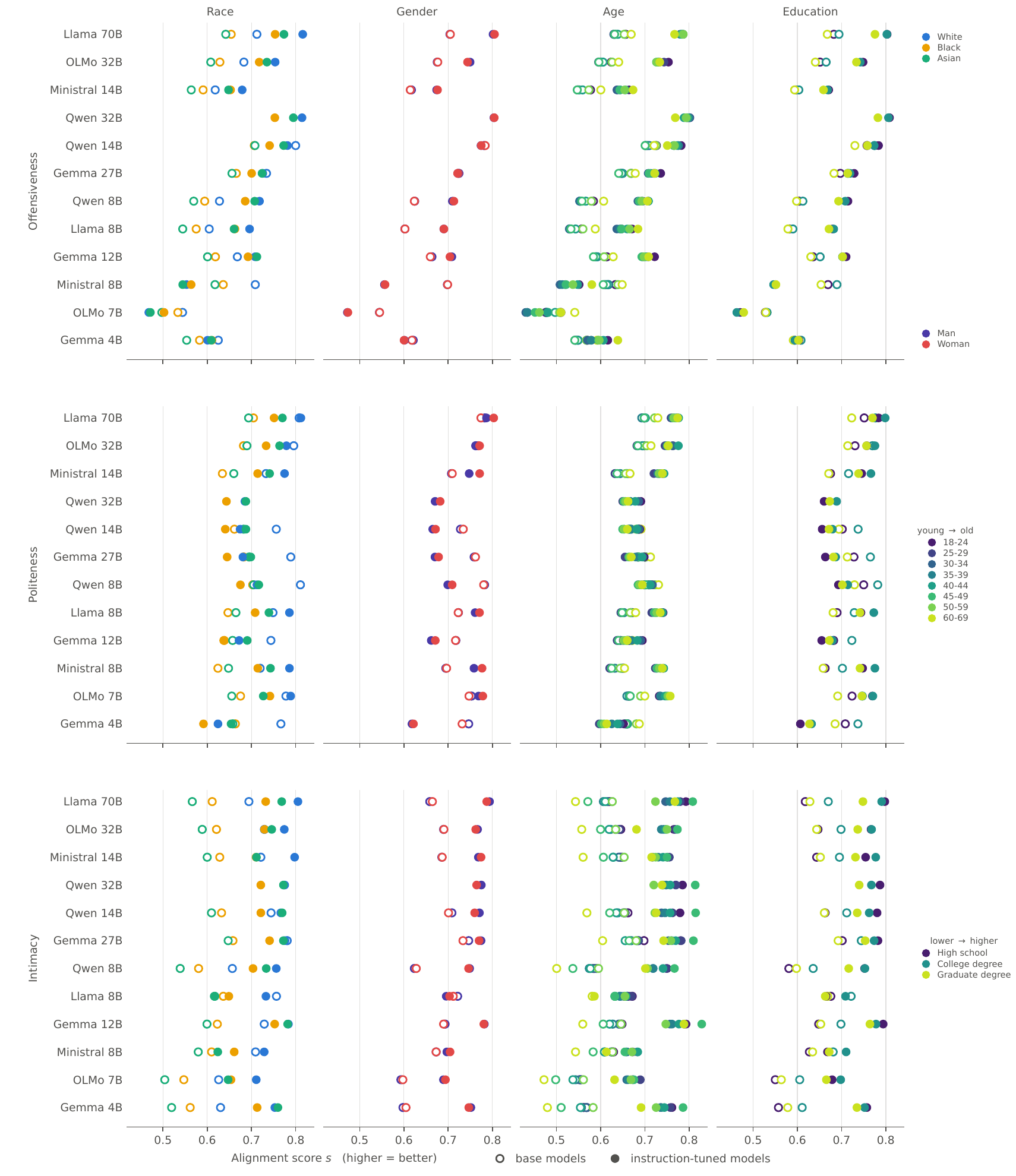}
\caption{Group alignment by task, one dot per demographic group.}
\label{fig:group-s-bytask}
\end{figure*}

This appendix reports the per-model, per-task view behind \S\ref{sec:results-default}: unconditioned predictions only, scored and macro-averaged as in \S\ref{sec:evaluation-metric}, over the retained groups. The \textit{base} pool barely moves across tasks (Figure~\ref{fig:group-s-bytask}): White is the closest race group and College degree the closest education band in every \textit{base} model on every task, and the oldest bracket leads in all but one model, except on intimacy, where the closest bracket splits between the youngest and 50--59. The \textit{instruct} pool varies more. White stays ahead of Black in all $12$ models on politeness and intimacy but in $9$ on offensiveness, where the mean gap drops to $+0.023$; Asian is the closest race group in half the models on politeness. Education reverses: College degree is closest in all $12$ models on politeness, the split is even on intimacy, and High school or below leads in $9$ of $12$ on offensiveness. Gender stays nearly flat on every task; age changes closest bracket from task to task.


\section{Per-Model Conditioning Effects}
\label{app:cond-tables}

\begin{figure}[t]
\centering
\includegraphics[width=\columnwidth]{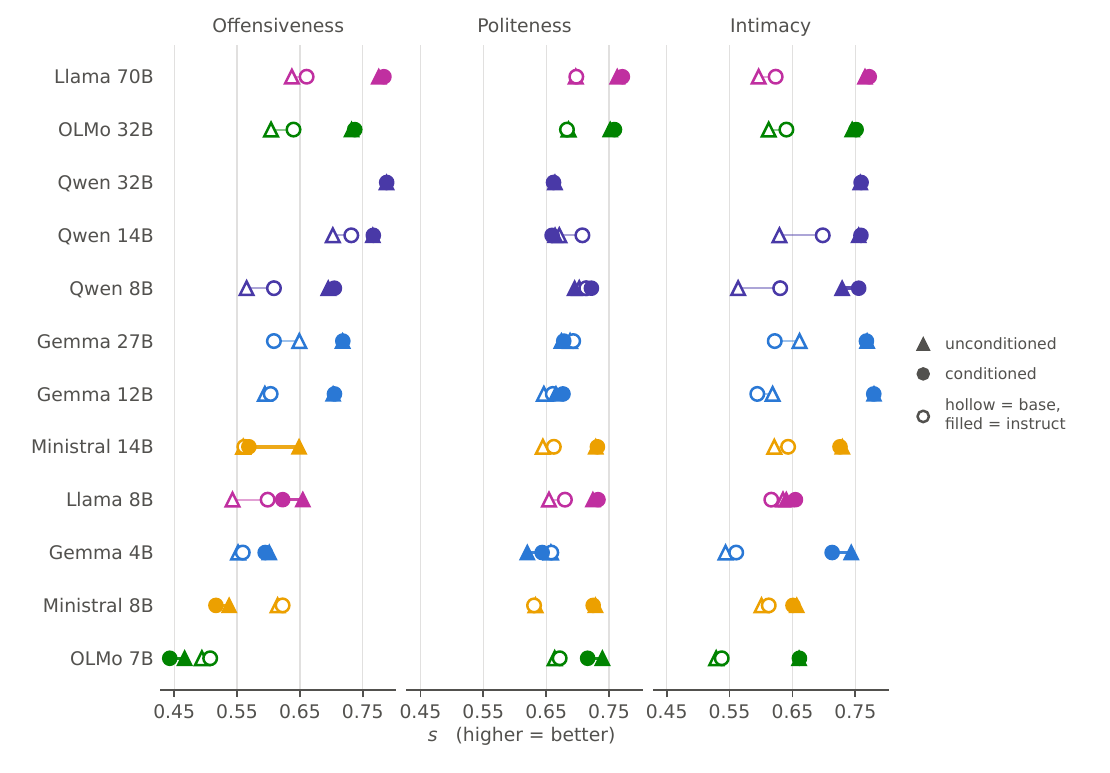}
\caption{Conditioning effects by task. 
Each row is one model.
The horizontal distance between the markers in a row is that model’s conditioning.
}
\label{fig:cond-bytask}
\end{figure}

This appendix reports the per-model, per-task view behind \S\ref{sec:results-conditioning}. 
Figure~\ref{fig:cond-bytask} splits the per-model view of Figure~\ref{fig:uncond-vs-cond} by
task, with rows in the same order. Two things are worth reading off directly: (i) on every task the hollow pairs shift right more often than the filled ones, and the \textit{base} mean is positive on all three tasks and exceeds the \textit{instruct} mean on all three; (ii) the tasks differ in dispersion rather than in direction. Politeness holds the smallest effects in both pools, while the largest sit on offensiveness for \textit{instruct} models and on intimacy for \textit{base} models. Offensiveness is the only task whose \textit{instruct} mean is negative, and since the pooled \textit{instruct} effect averages the three tasks, it is what places the pooled figure below zero. %

\section{Rating-Shift Faithfulness for Intersectional Profiles}
\label{app:faithfulness-intersect}


Table \ref{tab:shift-vs-dev-intersect} repeats the analysis of \ref{tab:shift-vs-dev} for the six gender$\times$race profiles.
Each profile shifts in the direction of its race attribute, with magnitudes attenuated toward the gender component. The reference deviations are computed from the annotators matching both attributes, so the intervals are wider than
in the single-attribute figure.

\begin{table}[th!]
\centering
\footnotesize
\setlength{\tabcolsep}{5pt}
\begin{tabular}{@{}l c c c@{}}
\textbf{Group} & \textbf{Human} & $\Delta$ \textbf{base} & $\Delta$ \textbf{instruct} \\
\toprule
\multicolumn{4}{@{}l}{\textit{Offensiveness}}\\
Man $\times$ White & $-0.008$ & $-0.068$ & $-0.025$ \\
Woman $\times$ White & $-0.019$ & $-0.049$ & $+0.005$ \\
Man $\times$ Asian & $+0.016$ & $-0.041$ & $+0.001$ \\
Woman $\times$ Asian & $-0.010$ & $-0.018$ & $+0.043$ \\
Man $\times$ Black & $+0.061$ & $-0.011$ & $+0.063$ \\
Woman $\times$ Black & $+0.061$ & $+0.011$ & $+0.104$ \\
\addlinespace
\multicolumn{4}{@{}l}{\textit{Politeness}}\\
Man $\times$ White & $+0.000$ & $+0.048$ & $+0.039$ \\
Woman $\times$ White & $+0.003$ & $+0.046$ & $+0.032$ \\
Man $\times$ Asian & $-0.027$ & $+0.041$ & $+0.026$ \\
Woman $\times$ Asian & $-0.030$ & $+0.037$ & $+0.011$ \\
Man $\times$ Black & $+0.050$ & $+0.023$ & $+0.005$ \\
Woman $\times$ Black & $+0.016$ & $+0.020$ & $-0.011$ \\
\addlinespace
\multicolumn{4}{@{}l}{\textit{Intimacy}}\\
Man $\times$ White & $-0.019$ & $-0.054$ & $-0.028$ \\
Woman $\times$ White & $+0.009$ & $-0.044$ & $-0.018$ \\
Man $\times$ Asian & $-0.045$ & $-0.047$ & $-0.018$ \\
Woman $\times$ Asian & $-0.023$ & $-0.039$ & $-0.004$ \\
Man $\times$ Black & $+0.037$ & $-0.038$ & $+0.011$ \\
Woman $\times$ Black & $+0.060$ & $-0.030$ & $+0.018$ \\
\bottomrule
\end{tabular}
\caption{Group deviation from the pooled human mean and the rating shift induced by conditioning for all models ($\Delta$), by task and gender$\times$race profile (mean rating on the [0,1] scale described in \S\ref{sec:evaluation-metric}). Faithful conditioning would give $\Delta\approx$~Human.}
\label{tab:shift-vs-dev-intersect}
\end{table}


\section{Controls for Distributional Spread}
\label{app:robustness-checks}


Scoring predictions 
introduces a sensitivity that a point-estimate comparison never faces: both sides of the comparison can be inflated by spread alone. A reference distribution built from many annotators covers several scale points, becoming a wide target that almost any prediction lands close to; a spread-out prediction, in turn, overlaps part of any reference by construction. A judge can therefore score higher on a group without simulating it any better. Because this risk is specific to distributional scoring, we introduce two controls that isolate and remove it: density matching equalizes the human reference, and a mode-accuracy reading equalizes the prediction. Effects that survive both are properties of the judge, not of distribution width.
A final check (\S\ref{app:token-coverage}) confirms that the models are placing probability mass on tokens relevant to the question rather than elsewhere.

\subsection{Density Matching}
\label{app:density-matching}



\begin{figure}[th!]
\centering
\includegraphics[width=\columnwidth]{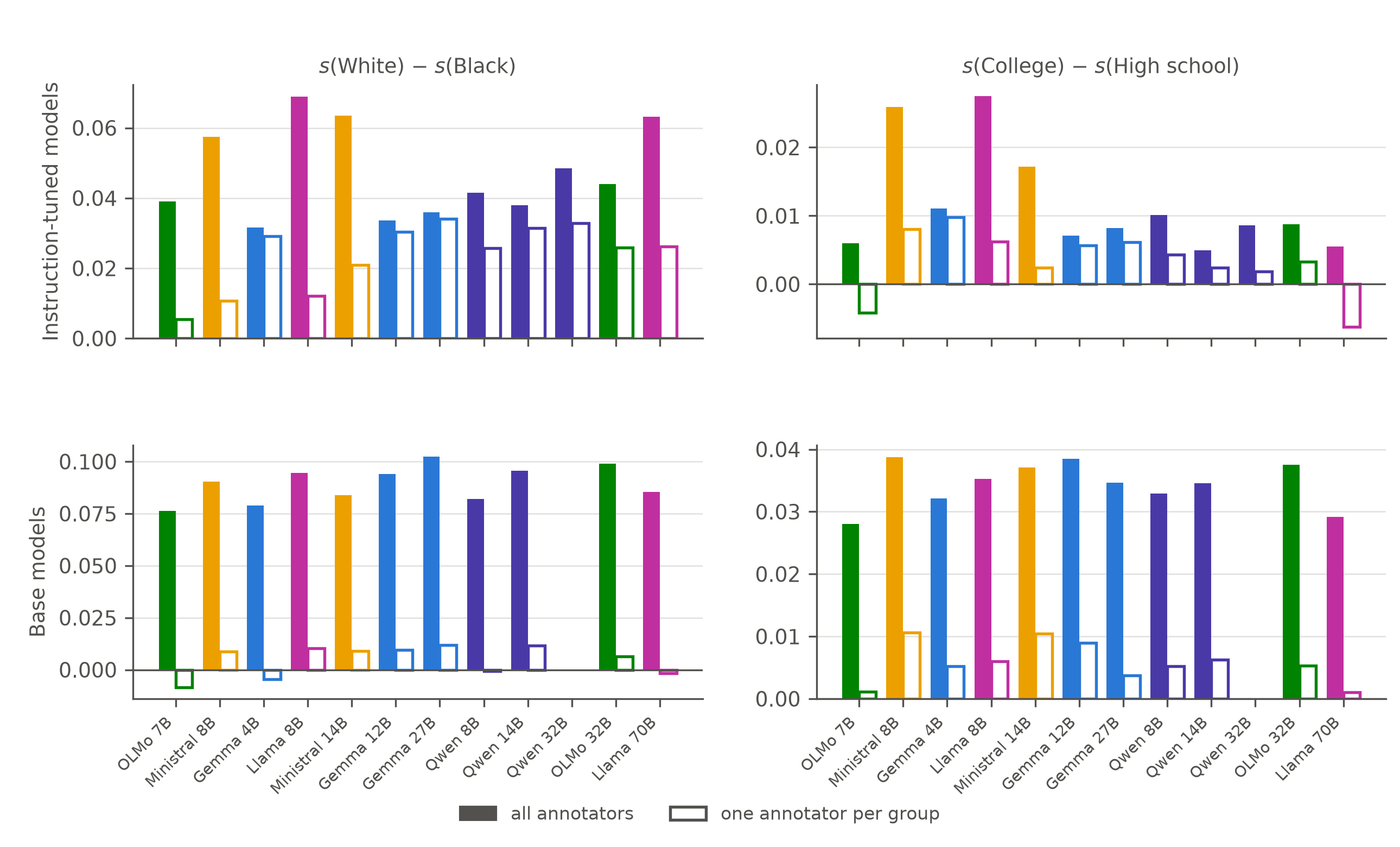}
\caption{Density matching per model (averaged over $20$ draws in filled bars). 
}
\label{fig:density-matching-permodel}
\end{figure}

Groups differ in how many annotators rate each item. A reference built from five annotators spreads over several scale points; one built from a single annotator is a point, which a prediction either matches or misses.
To measure how much of the alignment gaps this width difference explains, we replay every comparison with references of equal width: for each item we retain one random annotation per group, rebuild the references from those single labels, and recompute the gaps, averaging over $20$ random draws.
We match at one annotation because many Black and Asian cells contain only one.
Under matching (Figure~\ref{fig:density-matching-permodel}), the \textit{base}-pool gaps shrink substantially and a few change sign, indicating that most of the \textit{base} pool's raw gaps reflect annotator counts rather than judge behavior. The \textit{instruct} gaps also shrink but remain positive for race in every model, and reverse for education in only two. Gap \emph{magnitudes} are therefore partly a property of the data, whereas the \textit{instruct} pool's \emph{orderings} are a property of the judges; this is why \S\ref{sec:results-default} reports the orderings but does not interpret their sizes. 
The control thus separates a data artifact from a model effect that a single-target comparison could not have told apart.



\subsection{Mode Accuracy}
\label{app:top1}


The second control removes the corresponding advantage on the prediction side. We recompute every comparison under \emph{mode accuracy}: on each item the judge is credited only when its single most probable option is also the most frequent human label (ties included), on the same items and estimator as before. Because this reading compares only the top choice on each side, a spread-out prediction gains nothing from its spread.

\begin{figure}[ht!]
\centering
\includegraphics[width=\columnwidth]{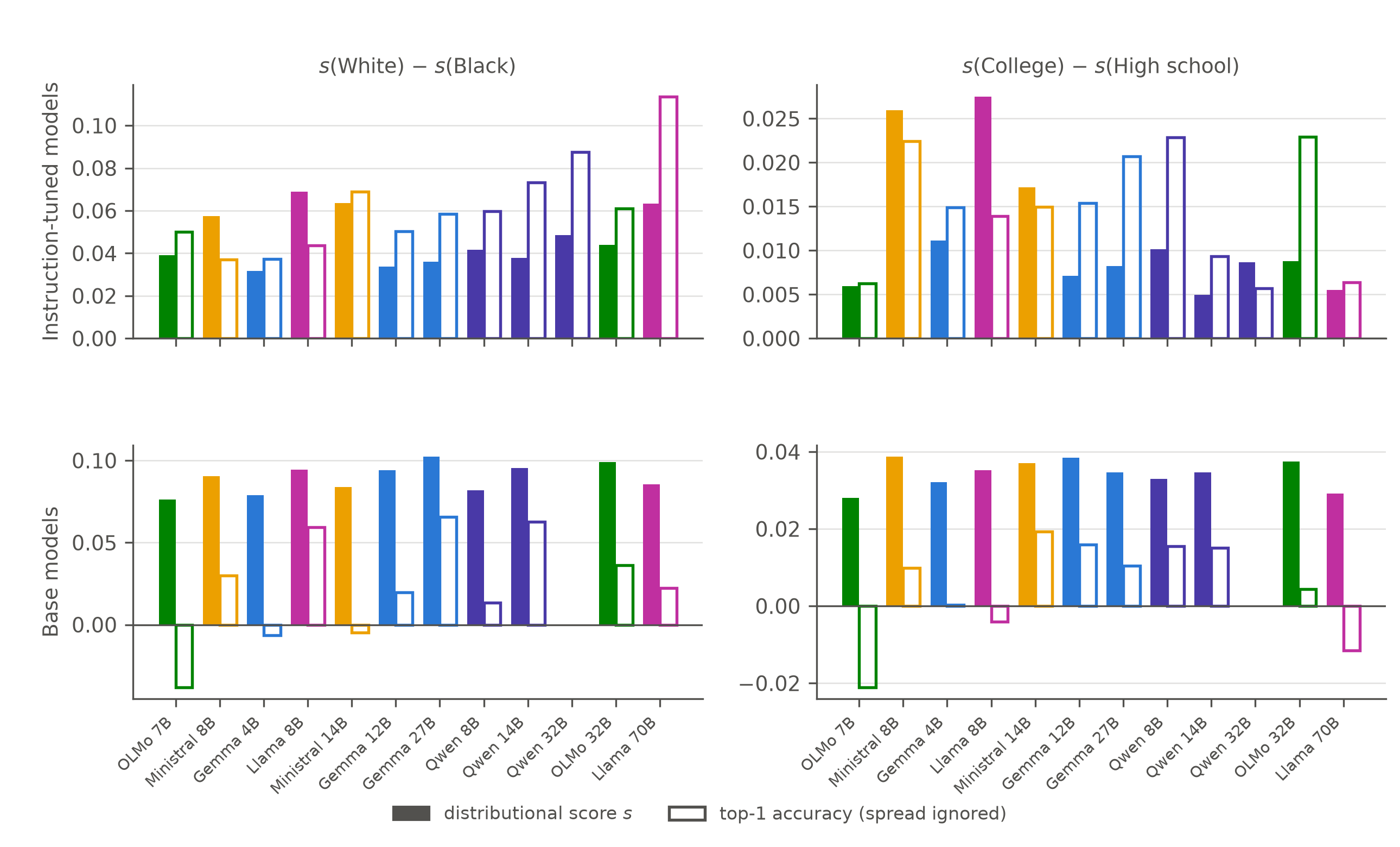}
\caption{Group gaps under the two readings. 
}
\label{fig:top1-contrast-permodel}
\end{figure}

%
Figure~\ref{fig:top1-contrast-permodel} compares the group gaps under the two readings, and the pattern mirrors density matching. In the \textit{base} pool the gaps largely close and a few change sign, confirming that most \textit{base}-pool gaps come from spread. In the \textit{instruct} pool the gaps do not close; for race they are mostly larger under mode accuracy than under $s$, and every \textit{instruct} model remains positive under both readings. The \textit{instruct} gaps therefore do not rest on distribution width.


Mode accuracy also tests whether the \textit{base}-model conditioning gain of \S\ref{sec:results-conditioning} is an artifact of near-flat predictions. 
It is not: under mode accuracy the gain 
doubles, from $+0.014$ to $+0.032$, and stays positive in 10 of 11 \textit{base} models, so conditioning changes which option a \textit{base} model selects, toward the option the group 
chose. The \textit{instruct} effect remains near zero under both readings. The per-group result of \S\ref{sec:results-groups} likewise holds: 14 of the 16 retained groups keep their sign, 
except for College degree, whose gain appears only under $s$. 
The two controls agree at the model level: both leave every \textit{instruct} contrast standing and attenuate the \textit{base} contrasts, with only 
Gemma~4B, OLMo~7B
failing both.
We also confirm that the conditioning effects are not driven by reference noise: we resample annotators within cells jointly with judges ($2{,}000$ replicates). The effects are mostly unchanged (Black $-0.033$, $[-0.048, -0.020]$; Woman$\times$Black on offensiveness $-0.087$, $[-0.132, -0.051]$), and single-annotator cells, which contribute no reference noise, are already covered by the density matching.

\subsection{Answer-Token Coverage}
\label{app:token-coverage}

Finally, we verify that the predicted distribution reflects a genuine judgment rather than an evasion. Because $p$ is a softmax over the five option tokens, a model that declined to answer would still yield a distribution over the scale. In practice the option tokens carry at least $96.7\%$ of the next-token mass in every \textit{base} model and $99.7\%$ in every instruction-tuned one, and on offensiveness, where refusal is likeliest, adding a demographic profile shifts that share by under $10^{-4}$. The effects we report are therefore changes in how the models judge, not changes in whether they answer.

\end{document}